\documentclass[journal]{IEEEtran}
\IEEEoverridecommandlockouts
\usepackage{amsmath,amssymb,amsfonts}
\usepackage{algorithm}
\usepackage{algpseudocode}
\usepackage{graphicx}
\usepackage{textcomp}
\usepackage{xcolor}
\usepackage{enumitem}
\usepackage{romannum}
\usepackage{gensymb}
\usepackage{multirow}
\usepackage{tabularx}
\usepackage{url}

\def\BibTeX{{\rm B\kern-.05em{\sc i\kern-.025em b}\kern-.08em
    T\kern-.1667em\lower.7ex\hbox{E}\kern-.125emX}}
\begin{document}

\title{Compliant Control of Quadruped Robots for Assistive Load Carrying \\
}


\author{
Nimesh Khandelwal, Amritanshu Manu,\\ Shakti S. Gupta, Mangal Kothari, Prashanth Krishnamurthy, Farshad Khorrami
}




\maketitle

\begin{abstract}

This paper presents a novel method for assistive load carrying using quadruped robots. The controller uses proprioceptive sensor data to estimate external base wrench, that is used for precise control of the robot's acceleration during payload transport. The acceleration is controlled using a combination of admittance control and Control Barrier Function (CBF) based quadratic program (QP). The proposed controller rejects disturbances and maintains consistent performance under varying load conditions. Additionally, the built-in CBF guarantees collision avoidance with the collaborative agent in front of the robot. The efficacy of the overall controller is shown by its implementation on the physical hardware as well as numerical simulations. The proposed control framework aims to enhance the quadruped robot's ability to perform assistive tasks in various scenarios, from industrial applications to search and rescue operations.

\end{abstract}


\section{Introduction}
The development of quadruped robots has expanded the frontiers of robotic mobility, particularly in navigating complex terrains where stability and adaptability are crucial. A key challenge for these robots is handling external disturbances, which demands advanced strategies to maintain balance during operation. This capability becomes important in collaborative scenarios - whether working alongside other robots or humans, especially for assistive load carrying. In such cases, the robot must modify its behavior in response to the externally applied forces while maintaining stability to execute collaborative tasks effectively.

Works by \cite{kim2023layered, yang2022collaborative} explored collaborative quadruped control for load-carrying applications, using layered planning with predictive approaches for towing via cables and rods. \cite{liu2024optimization} proposed flocking control and gait synchronization using optimization, while \cite{sombolestan2023hierarchical} demonstrated collaborative object manipulation using quadruped robots by hierarchical adaptive controller with a decentralized loco-manipulation controller.

Admittance control enhances the dynamic response of quadruped robots by adjusting movements based on external forces, enabling compliant behavior crucial for disturbance rejection. Since external disturbances vary with the environment, estimating these forces is necessary, either through direct sensing or indirect estimation. Various approaches have been explored for designing admittance/impedance controllers for legged robots. One prominent method in the literature is to design an impedance controller in the robot's foot task space (referred to as the operational space in this paper). \cite{focchi2012torque} implemented an end-effector Cartesian impedance controller to generate compliant motions for the hydraulic HyQ robot \cite{semini2010design}, building on their earlier work \cite{ugurlu2013dynamic} that used active compliance. \cite{park2012impedance} employed a similar foot impedance controller to modulate trotting behavior on uneven terrain, while \cite{buchli2009compliant} used active compliance for robust locomotion on rough surfaces. \cite{pedro2024quadruped} followed a similar approach for a quadruped robot using a differential-based planar motion generation strategy. All these studies focus on foot impedance control to enhance robustness on uneven terrain but do not address base motion modulation or the estimation of external forces on the robot's base.

None of the works mentioned above discuss the requirement of compliant base motion for stability when external forces act on the robot base. Recently, \cite{lee2022deep} employed reinforcement learning (RL) to train a locomotion policy that enables compliant base motion under external disturbances, improving robustness, energy efficiency, and safety. This was further extended by \cite{hartmann2024deep} to design RL-based controllers for compliant motion in quadrupeds. In contrast, our work adopts a model-based approach to design a controller that generates compliant base motion for a quadruped using admittance control. To estimate the external base wrench, \cite{kang2023view} used multiple sensing modalities. In our work, the external base wrench is estimated using only the robot’s proprioceptive information, eliminating the need for additional sensors, which enhances efficiency and reliability by avoiding potential points of failure.

While performing collaborative tasks, it is important to avoid collisions with other agents in the environment. \cite{kim2023safety} used a decentralized approach similar to \cite{kim2023layered}, incorporating CBF constraints in an optimization-based framework. \cite{unlu2024control} used CBF with circulation inequality to generate safe paths by providing high-level velocity commands. Both approaches rely on SLAM to evaluate CBF constraints. \cite{hao2024d2slam} demonstrated collaborative SLAM using near-field estimation for precise state estimation. A similar approach for collaborative multi-UAV exploration was proposed by \cite{boyu2023racer} using an online hgrid space decomposition based strategy. A more direct method using point-cloud-based CBFs for drone control was proposed by \cite{de2024point} to reduce computational demands. We employ a similar approach, utilizing point-cloud data from the robot's LiDAR sensor. Note that we focus on collision avoidance with a collaborative agent in front of the robot, rather than a general safe planning problem since our focus is on the control design for assistive load carrying.

This research contributes to the field of model-based collaborative control of legged robots. The design of admittance control for quadruped robot using only proprioceptive data is one of the novel contributions. Another contribution is the kinematic base motion planning where a single acceleration command is used to achieve the three-fold objective of: 1) tracking user command, 2) achieving compliant motion, and 3) avoiding collisions without any online optimization. Our approach differs from existing work by decoupling the control framework rather than using computationally expensive single-optimization solutions. This results in a simplified, efficient system that integrates these methodologies in a novel way. Accuracy of the proprioceptive force estimation is shown experimentally. Two use cases of such controller design are discussed, namely: 1) disturbance rejection, and 2) collaborative load carrying, the results for which are shown experimentally. Separate experiments for human-robot and robot-robot collaborative load carrying are performed to show the invariance of the controller to the source of external forcing. The results from this study are also helpful for designing RL-based policies to include compliance appropriately in the reward structure during the training process. 

\section{PROPOSED APPROACH}
\subsection{Problem formulation and terminology}
This paper addresses the problem of designing a controller for the quadruped robot to perform compliant motion in response to the external force applied on the robot base by a \textit{leader}, while maintaining stability on a flat terrain. The \textit{leader} can be a human or another robot. The robot that moves in response to the applied force is termed the \textit{follower} robot. The controller design is done for the \textit{follower} robot, which in this case is a quadruped robot with 18 total Degrees of Freedom (DOF), namely Unitree Go2. These are represented by the generalized coordinate vector $\mathbf{q}\in\mathbb{R}^{18}$. The DOF are partitioned into the 6-DOF unactuated space for the robot base and the 12-DOF actuated space for the leg joints with 3 joints in each leg, as $\mathbf{q}=\begin{bmatrix}
    \mathbf{q}_b^T&\mathbf{q}_j^T
\end{bmatrix}^T$. $\mathbf{q}_b=\begin{bmatrix}
    \mathbf{r}_b^T&\mathbf{\Phi}_b^T
\end{bmatrix}^T\in \mathbb{R}^6$, where $\mathbf{r}_b\in\mathbb{R}^{3}$ is the position vector of the robot center-of-mass (COM) and $\mathbf{\Phi}_b\in \mathbb{R}^3$ is the \textit{XYZ}-Euler angle vector representing the orientation of the robot base. $\mathbf{q}_j\in\mathbb{R}^{12}$ is the vector of joint angles for all four legs. The legs are ordered as front-left (FL), front-right (FR), rear-left (RL), and rear-right (RR), with each leg numbered 0, 1, 2, and 3, respectively.
\subsection{Mathematical model}
The general equation of motion (EOM) of a quadruped robot, modeled as a rigid-body system with contact constraints, is expressed as:
\begin{equation}
    \mathbf{M}(\mathbf{q})\ddot{\mathbf{q}} + \boldsymbol{\eta}(\mathbf{q}, \dot{\mathbf{q}}) = \mathbf{S}^T\boldsymbol{\tau} + \mathbf{J}_c^T\mathbf{F}_c
    \label{eq:eom}
\end{equation}
where $\mathbf{M}(\mathbf{q})\in\mathbb{R}^{18\times 18}$ is the joint space inertia matrix of the system, and $\boldsymbol{\eta}(\mathbf{q}, \dot{\mathbf{q}})\in\mathbb{R}^{18}$ represents the nonlinear terms corresponding to coriolis, centrifugal and gravitational effects. $\mathbf{S} = \begin{bmatrix}
    \mathbf{0}_{12\times6}&\mathbf{I}_{12\times 12}
\end{bmatrix}$ is the selection matrix representing the underactuation since only the joint DOFs are actuated. $\boldsymbol{\tau}\in\mathbb{R}^{12}$ is the joint torque command, $\mathbf{J}_c\in\mathbb{R}^{12\times 18}$ is the contact Jacobian matrix, and $\mathbf{F}_c\in\mathbb{R}^{12}$ are the contact forces on the feet of the robot.
The frame notation used in this paper is similar to \cite{khandelwal2024distributed}. Any differences are mentioned explicitly.

We also define the operational-space coordinates as
$\boldsymbol{\chi} = \begin{bmatrix}
\mathbf{q}_b^T&&\mathbf{r}_{p_1}^T&&\mathbf{r}_{p_2}^T&&\mathbf{r}_{p_3}^T&&\mathbf{r}_{p_4}^T\end{bmatrix}^T\in\mathbb{R}^{18\times 1}$, where $\mathbf{r}_{p_i}\in\mathbb{R}^3(\forall{i\in\{1,2,3,4}\})$ is the position of foot $i$ in the global coordinate system. In Eq. (\ref{eq:eom}), there is an implicit assumption that only the feet of the robot are subjected to external forces. If there are external forces applied to the robot base as well, it should be modified to
\begin{equation}
    \begin{aligned}
        \mathbf{M}(\mathbf{q})\ddot{\mathbf{q}} + \boldsymbol{\eta}(\mathbf{q}, \dot{\mathbf{q}}) = \mathbf{S}^T\boldsymbol{\tau} + \mathbf{J}^T(\mathbf{q})\mathbf{F}_e
    \label{eq:eom2}
    \end{aligned}
\end{equation}
where $\mathbf{F}_e\in\mathbb{R}^{18}$ are the generalized external forces on the system being applied on all the operational space DoFs. These are the external forces that we will be estimating. $\mathbf{J}(\mathbf{q})\in\mathbb{R}^{18\times 18}$ is the Jacobian of the end-effector coordinates ($\boldsymbol{\chi}$) w.r.t. the generalized coordinates ($\mathbf{q}$). It is defined as:
\begin{equation}
    \begin{aligned}
        \mathbf{J}(\mathbf{q})=\frac{\partial \boldsymbol{\chi}}{\partial \mathbf{q}}.
    \end{aligned}
\end{equation}
The rate of change of joint space coordinates and the end-effector coordinates are related by this Jacobian as:
\begin{equation}
    \begin{aligned}
        \dot{\boldsymbol{\chi}}&=\mathbf{J}(\mathbf{q})\dot{\mathbf{q}}\\
        \ddot{\boldsymbol{\chi}}&=\mathbf{J}(\mathbf{q})\ddot{\mathbf{q}} + \dot{\mathbf{J}}(\mathbf{q}, \dot{\mathbf{q}})\dot{\mathbf{q}}
    \end{aligned}
    \label{eq:ee_acceleration}
\end{equation}
where $\dot{\boldsymbol{\chi}}$ and $\ddot{\boldsymbol{\chi}}$ are the velocity and acceleration of the end-effector coordinates. The procedure to estimate the $\mathbf{F}_e$ is discussed in the Section \ref{sec:base_wrench}.
\subsection{Base wrench estimation}
\label{sec:base_wrench}
To estimate the external wrench being applied on the robot base, we use the available proprioceptive data: joint angles ($\mathbf{q}_j)$, joint velocities ($\dot{\mathbf{q}}_j$), and joint torques ($\boldsymbol{\tau}$). These values are used in the state-estimator as described in \cite{cheetah3} to get $\hat{\mathbf{q}}=\begin{bmatrix}
    \hat{\mathbf{q}}_b^T&\hat{\mathbf{q}}_j^T
\end{bmatrix}^T$ and $\hat{\dot{\mathbf{q}}}=\begin{bmatrix}
    \hat{\dot{\mathbf{q}}}_b^T&\hat{\dot{\mathbf{q}}}_j^T
\end{bmatrix}^T$, which are the estimated generalized coordinate and velocity vector. We use the following relation to estimate the external forces on all the operational space DOF:
\begin{equation}
    \begin{aligned}
        \hat{\mathbf{F}}_e=-(\mathbf{J}(\hat{\mathbf{q}})^T)^{\dagger}\left(\mathbf{S}^T\hat{\boldsymbol{\tau}} - \boldsymbol{\eta}(\hat{\mathbf{q}}, \hat{\dot{\mathbf{q}}})\right)
    \end{aligned}
\end{equation}
where $\hat{\mathbf{F}}_e$ is the vector of estimated forces in the end-effector coordinates. The $(\hat.)$ represents the estimated/filtered value of the mentioned quantity. The proprioceptive data used here, i.e., $\hat{\mathbf{q}}_j$, $\hat{\dot{\mathbf{q}}}_j$, and $\hat{\boldsymbol{\tau}}$, are filtered values of the raw sensor data obtained from the joint encoder and the motor controller. For filtering of the sensor data, a second order Butterworth filter was used. We neglect the effects of the inertial terms on the estimated force vector. The estimated force vector can be written as: $\hat{\mathbf{F}}_e=\begin{bmatrix}
    \hat{\mathbf{F}}_b^T&\hat{\mathbf{F}}_c^T
\end{bmatrix}^T$, where $\hat{\mathbf{F}}_b\in\mathbb{R}^6$ is the estimated external wrench on the robot base. This wrench is used in the admittance controller as shown in Section \ref{sec:admittance}. The estimated foot contact forces, $\hat{\mathbf{F}}_c\in\mathbb{R}^{12}$ are used in the contact detection algorithm as described in \cite{bledt2018contact}.

All the quantities in this paper are expressed either in the global (inertial) frame $\mathcal{G}$ or the local frame  $\mathcal{B}$ that is attached to the COM and has the same orientation as the robot base. The axes of these frames are denoted with a superscript describing the frame, e.g. $X^{\mathcal{G}}$, whereas the notation $^{\mathcal{B}}_{}\mathbf{r}_b$ implies that $\mathbf{r}_b$ is expressed in the frame $\mathcal{B}$.
\subsection{Base motion generation}
\label{sec:base_motion}
To generate the nominal reference trajectory of the robot base, we use a feedback based acceleration limited trajectory generation. The input to the robot is assumed to be the desired velocity commands $^\mathcal{B}\mathbf{v}_{cmd}=\begin{bmatrix}
    v_x&v_y&\omega_z
\end{bmatrix}^T$, which is used to calculate the desired acceleration of the robot as discussed in \ref{sec:admittance}. To generate nominal base reference motion in the global $xy$-plane, :
\begin{equation}
    \begin{aligned}
        ^\mathcal{G}\dot{\mathbf{r}}_{b,d,k}^{x,y}&=\min\{\max\{^\mathcal{G}\dot{\mathbf{r}}_{b,d,k-1}^{x,y}+^\mathcal{G}\ddot{\mathbf{r}}_{b,d,k}^{x,y}\Delta t, -\mathbf{v}_{max}\}, \mathbf{v}_{max}\}\\
        ^\mathcal{G}\mathbf{r}_{b,d,k}^{x,y}&=^\mathcal{G}\mathbf{r}_{b,d,k-1}^{x,y}+^\mathcal{G}\dot{\mathbf{r}}_{b,d,k}^{x,y}\Delta t+0.5^\mathcal{G}\ddot{\mathbf{r}}_{b,d,k}^{x,y}\Delta t^2
    \end{aligned}
    \label{eq:velocity_update}
\end{equation}
where $\mathbf{r}_{b,d,k}$, $\dot{\mathbf{r}}_{b,d,k}$, and $\ddot{\mathbf{r}}_{b,d,k}$ represent the desired position, velocity, and acceleration of the base at the $k$-th iteration. $\Delta t$ is the time difference between current and previous iteration. $\mathbf{v}_{max}\in\mathbb{R}^2$ is the vector of maximum allowed base velocities in the horizontal plane. They are calculated based on the current robot configuration and the tracking controller being used. The calculation of base acceleration used for the position and velocity update is given in Sections \ref{sec:admittance} and \ref{sec:cbf}.
\subsection{Admittance control on robot base}
\label{sec:admittance}
To generate the nominal base trajectories discussed in Section \ref{sec:base_motion}, the acceleration of the base is calculated based on the user input. The estimated external base wrench is used to design the proposed admittance controller. The commanded acceleration of the robot, $^\mathcal{B}\ddot{\mathbf{r}}_{b,d,k'}^{x,y}$, is calculated as:
\begin{equation}
    \begin{aligned}
        ^\mathcal{B}\ddot{\mathbf{r}}_{b,d,k'}^{x,y} &= K_b(^\mathcal{B}\mathbf{v}_{cmd,k-1}^{x,y} - ^\mathcal{B}\dot{\mathbf{r}}_{b,d,k-1}^{x,y}) + K_f(^\mathcal{B}\hat{\mathbf{F}}_{b,k}^{x,y})\\
    \end{aligned}
    \label{eq:acc_admittance}
\end{equation}
where $\hat{\mathbf{F}}_{b,k}^{x,y}$ is the estimated force on the robot base in the global $xy$-plane at the $k$-th time-step. The $k'$ index indicates that this quantity is not the final value used for base motion update, and will be updated in later sections. $K_b$ is the proportionality constant relating the desired base acceleration to the difference in user velocity command and generated desired base velocity. $K_f$ is the constant determining the effect of applied base force on the robot base acceleration. To determine the values of constants $K_b$ and $K_f$, we use the maximum value constraints on the evolution trajectory of the robot base. 
For simplifying the analysis, we assume $\mathbf{v}_{cmd}=\begin{bmatrix}
    0&0&0
\end{bmatrix}^T$ (pure push/pull of the robot base), $0\le\Delta t\ll1$, and ignore any saturation. Under these assumptions, Eq. (\ref{eq:acc_admittance}) can be re-written in the following form:
\begin{equation}
    \begin{aligned}
        &\ddot{\mathbf{r}}_{b,d,k}^{x,y}+K_b\dot{\mathbf{r}}_{b,d,k}^{x,y}=K_f\hat{\mathbf{F}}_{b,k}^{x,y}\\
    \Rightarrow&\dot{\mathbf{v}}+K_b\mathbf{v}=\mathbf{F}_{ext}
    \end{aligned}
\end{equation}
where $\mathbf{v}=\dot{\mathbf{r}}_{b,d,k}^{x,y}$, $\mathbf{F}_{ext}=K_f\hat{\mathbf{F}}_{b,k}^{x,y}$. Since this is a first order system, its time evolution can be analytically calculated (assuming $\mathbf{v}(0)=0$ and $\mathbf{F}_{ext}$ is constant):
\begin{align*}
    \mathbf{v} &= \frac{\mathbf{F}_{ext}}{K_b}\left(1 - e^{-K_bt}\right)\\
    \dot{\mathbf{v}}&=\mathbf{F}_{ext}e^{-K_bt}.
\end{align*}
Considering the maximum desired acceleration and velocity as $\mathbf{a}_{max}$ and $\mathbf{v}_{max}$
, respectively, the constants $K_f$ and $K_b$ can be calculated as:
\begin{equation}
    \begin{aligned}
        \mathbf{a}_{max}&=\dot{\mathbf{v}}_{max}=K_f|\hat{\mathbf{F}}_{b,k}^{x,y}|_{max}\Rightarrow K_f=\frac{\mathbf{a}_{max}}{|\hat{\mathbf{F}}_{b,k}^{x,y}|_{max}},\\
    \mathbf{v}_{max}&=\frac{K_f|\hat{\mathbf{F}}_{b,k}^{x,y}|_{max}}{K_b}\Rightarrow K_b=\frac{\mathbf{a}_{max}}{\mathbf{v}_{max}}
    \end{aligned}
\end{equation}
where $|\hat{\mathbf{F}}_{b,k}^{x,y}|_{max}$ is the theoretical maximum estimated horizontal force on the base of the robot. This value is calculated using the estimated maximum total normal force on the feet of the robot and the friction coefficient.

\subsection{CBF for collision avoidance}
\label{sec:cbf}
During collaborative load carrying, it is desirable to avoid collisions with the collaborative agent if it comes close to the other robot. To achieve this, we can calculate additional desired base acceleration for collision avoidance using Control Barrier Functions (CBFs). This is enabled by the simplified base motion generation pipeline, that allows us to calculate multiple acceleration commands for multiple objectives. To further simplify the formulation, we design a 1-D CBF along the line connecting the robot base and nearest coordinates on the collaborative agent. We use the leader-follower CBF formulation used by \cite{ames2016control} for Adaptive Cruise Control. We modify the definition of the problem to fit our scenario: 1) the leader is the agent that is controlling or guiding the follower robot, and 2) the system has unit mass and no friction. The dynamic equations for this system can be written as:
\begin{equation}
    \begin{aligned}
        \begin{bmatrix}
            \dot{r}_b\\\dot{v}_b
        \end{bmatrix}&=\begin{bmatrix}
            v_b\\0
        \end{bmatrix}+\begin{bmatrix}
            0\\1
        \end{bmatrix}a_b\\
        \Rightarrow\dot{\mathbf{y}}&=\mathbf{f}(\mathbf{y})+\mathbf{g}(\mathbf{y})a_b
    \end{aligned}
    \label{eq:cbf_system}
\end{equation}
where $\mathbf{y}=\begin{bmatrix}
    r_b,v_b
\end{bmatrix}^T$ is the state vector, $r_b\in\mathbb{R}$ is the distance between the robot base, $\hat{\mathbf{r}}_{b}^{x,y}$, and the nearest coordinates of the leader, $\hat{\mathbf{r}}_{lead}^{x,y}$, along the line connecting the two, $\hat{\mathbf{o}}$. $v_b\in\mathbb{R}$ is the rate of change of $r_b$. $a_b$ is the acceleration applied to control the system in Eq. (\ref{eq:cbf_system}). The constraint for this system is that the robot should always maintain a distance $\delta_{CBF}$ from the leader. For the mentioned system, this is written as:
\begin{equation}
    \begin{aligned}
        h(z)\triangleq -r_b-\delta_{CBF} - T_hv_b\ge0
    \end{aligned}
\end{equation}
where $T_h$ is the look-ahead time required for deceleration. We use this to define the CBF as:
\begin{equation}
    \begin{aligned}
        H(\mathbf{y}) = -r_b - \delta_{CBF}-T_hv_b-\frac{1}{2}\frac{v_b^2}{a_{max}}.
    \end{aligned}
\end{equation}
The final CBF constraint is written as:
\begin{equation}
    \begin{aligned}
        \dot{H}(\mathbf{y})&\ge-\alpha H(\mathbf{y})\\
        \Rightarrow \left(\frac{\partial H}{\partial \mathbf{y}}\right)\frac{\partial \mathbf{y}}{\partial t}+\alpha H(\mathbf{y})&\ge 0\\
        \Rightarrow (L_\mathbf{f}H)+(L_\mathbf{g}H)a_b+\alpha H(\mathbf{y})&\ge 0
    \end{aligned}
\end{equation}
where $L_\mathbf{f}H$ and $L_\mathbf{g}H$ are the Lie derivatives of the CBF, $H(\mathbf{y})$, along $\mathbf{f}$ and $\mathbf{g}$, respectively. The optimal acceleration that satisfies this constraint is the solution of the CBF-QP:
\begin{equation}
    \begin{aligned}
        &a_b^*=\arg\min_{a_b}(a_b)^2\\
        &\text{subject to: } \dot{H}(\mathbf{y})+\alpha H(\mathbf{y})\ge0.
    \end{aligned}
\end{equation}
The closed form solution of this QP is:
\begin{equation}
    \begin{aligned}
        a_b^*=\begin{cases}
            0 &\text{ if  } \Gamma\ge0\\
            -\frac{\Gamma}{(L_\mathbf{g}H)} &\text{otherwise}
        \end{cases}
    \end{aligned}
    \label{eq:closed_form_qp}
\end{equation}
where $\Gamma=L_\mathbf{f}H+\alpha H$ is the CBF constraint evaluated at $a_b=0$. This simplification can be done here because all the individual terms in the equation are scalar.
For the system described in Eq. (\ref{eq:cbf_system}), the partial derivative of the CBF function with respect to $\mathbf{y}$ is:
\begin{equation}
    \begin{aligned}
        \frac{\partial H}{\partial \mathbf{y}}&=\begin{bmatrix}
            0&&-T_h-\frac{v}{a_{max}}
        \end{bmatrix}.
    \end{aligned}
    \label{eq:cbf_jac}
\end{equation}
Using Eq. (\ref{eq:cbf_jac}), we can calculate the required Lie derivatives:
\begin{equation}
    \begin{aligned}
        L_\mathbf{f}H&=\frac{\partial H}{\partial \mathbf{y}}\mathbf{f}(\mathbf{y})
        = -v_b,\\
        L_\mathbf{g}H&=\frac{\partial H}{\partial \mathbf{y}}\mathbf{g}(\mathbf{y})
        =-T_h-\frac{v_b}{a_{max}}.
    \end{aligned}
    \label{eq:lyapunov_values}
\end{equation}
The value of $\Gamma$ in Eq. (\ref{eq:closed_form_qp}) can be written as:
\begin{equation}
    \begin{aligned}
        \Gamma&=-v_b+\alpha\left(-r_b - \delta_{CBF}-T_hv_b-\frac{1}{2}\frac{v_b^2}{a_{max}}\right).
    \end{aligned}
\end{equation}

To get the nearest coordinates of the leader robot, $\hat{\mathbf{r}}_{lead}^{x,y}$, we use the LiDAR sensor mounted on the follower robot. The point cloud received from the robot is cropped using a box filter. The cropped point cloud is then processed to extract the nearest point to the follower robot base using K-Nearest Neighbour (KNN) \cite{fix1985discriminatory} search using Fast Library for Approximate Nearest Neighbors (FLANN) \cite{muja2009fast}. All the point cloud computations were done using the Point Cloud Library (PCL) library \cite{rusu20113d}. Since the LiDAR is rigidly attached to the robot base, the coordinates received from the KNN-search are in the robot base frame and do not need further transformations. The coordinates of the nearest point are used to evaluate the CBF constraint and then solve Eq. (\ref{eq:closed_form_qp}). Since there is no assumption on the shape of the leader, the proposed controller is able to avoid collisions with any generic leader that lies within the cropped box in front of the follower robot.
The final base acceleration to be used in Eq. (\ref{eq:velocity_update}) is:
\begin{equation}
    \begin{aligned}
        ^\mathcal{B}\ddot{\mathbf{r}}_{b,d,k{''}}^{x,y}&=^\mathcal{B}\ddot{\mathbf{r}}_{b,d,k'}^{x,y}+a_b^*  \; ^\mathcal{B}\hat{\mathbf{o}}\\
        ^\mathcal{B}\ddot{\mathbf{r}}_{b,d,k}^{x,y}&=\min\{\max\{^\mathcal{B}\ddot{\mathbf{r}}_{b,d,k''}^{x,y}, -\mathbf{a}_{max}\}, \mathbf{a}_{max}\}.
    \end{aligned}
\end{equation}

\section{FOOTSTEP PREDICTION AND MOTION CONTROL}
The swing foot placement algorithm is based on the Algorithm 2 described in \cite{khandelwal2024distributed}. To make the motion of the robot robust to external disturbances, the desired base linear velocity, $\dot{\mathbf{r}}_{b,d,k}^{x,y}$, is saturated to be within a ball of radius $\delta_v$ around the estimated base velocity, $\hat{\dot{\mathbf{r}}}_{b,k}^{x,y}$, as:
\begin{equation}
    \begin{aligned}
        \dot{\mathbf{r}}_{b,d,k}^{x,y}=\min(\max(\dot{\mathbf{r}}_{b,d,k}^{x,y}, \hat{\dot{\mathbf{r}}}_{b,k}^{x,y} - \delta_v), \hat{\dot{\mathbf{r}}}_{b,k}^{x,y} + \delta_v).
    \end{aligned}
\end{equation}
The footstep placement heuristic is also changed to use the actual hip velocity , $\hat{\dot{\mathbf{r}}}_{h_i,k}^{x,y}$, along with a scaled capture point heuristic. Line 4 in Algorithm 2 in \cite{khandelwal2024distributed} is changed to:
\begin{equation}
    \begin{aligned}
        \mathbf{d}_i^{x,y} = \frac{1}{2}\hat{\dot{\mathbf{r}}}_{h_{i,k}}^{x,y}\phi_{st,i}t_g+k_{cp}\sqrt{\frac{r_{h_i,d,k}^z}{||\mathbf{g}||}}\left(\hat{\dot{\mathbf{r}}}_{b,k}^{x,y} - \dot{\mathbf{r}}_{b,d,k}^{x,y}\right)
    \end{aligned}
\end{equation}
where $\mathbf{d}_i^{x,y}$ is the desired position of the $i$-th foot with respect to the corresponding desired hip position, $\mathbf{r}_{h_i,d,k}$, at the next touchdown. $k_{cp}$ is the scaling coefficient to modify the contribution of capture point heuristic to the overall foot placement. $\phi_{st,i}$ and $t_g$ are the stance fraction for the $i$-th feet and the gait period, respectively. $\mathbf{g}\in\mathbb{R}^3$ is the gravity vector. The final reference touchdown position for the $i$-th foot is:
\begin{equation}
    \begin{aligned}
        \mathbf{r}_{p_i,d,k}^{x,y}=\mathbf{r}_{h_i,d,k}+\mathbf{d}_i^{x,y}.
    \end{aligned}
\end{equation}
To track the reference velocity calculated by the planner, we use the hierarchical asynchronous reactive-predictive control architecture. The predictive controller is the reduced order ConvexMPC \cite{convex-mpc} that runs asynchronously on an independent thread and is given by:
\begin{equation}
    \begin{aligned}
        \min_{\mathbf{u}_j} \quad & \sum_{j=0}^{N}(\mathbf{x}_{ref,j}-\mathbf{x}_j)^T\mathbf{Q}(\mathbf{x}_{ref,j}-\mathbf{x}_j)\\
        &+(\mathbf{u}_{ref,j}-\mathbf{u}_j)^T\mathbf{R}(\mathbf{u}_{ref,j}-\mathbf{u}_j)\\
        \text{subject to} \quad &\mathbf{x}_{j+1}=\mathbf{A}_j\mathbf{x}_j+\mathbf{B}_j\mathbf{u}_j+\mathbf{b}_j\\
        &\mathbf{lb}_{i}\le\mathbf{D}\mathbf{u}_{j}^i\le\mathbf{ub}_{i}\ \ \ \forall i \in \{0,1,2,3\}
    \end{aligned}
    \label{eq:mpc_optimization}
\end{equation}
where $\mathbf{x} = \begin{bmatrix}
    \mathbf{r}_b^T&\boldsymbol{\Phi}_b^T&\dot{\mathbf{r}}_b^T&\boldsymbol{\omega}_b^T
\end{bmatrix}^T$ is the state vector for base motion and $\mathbf{u}=\begin{bmatrix}
    \mathbf{F}_1^T&\mathbf{F}_2^T&\mathbf{F}_3^T&\mathbf{F}_4^T
\end{bmatrix}^T$ is the vector of contact forces on the robot feet. $N$ is the total number of nodes in the MPC problem, where each node is denoted by index $j$. $\mathbf{Q}\in\mathbb{R}^{12\times 12}$ and $\mathbf{R}\in\mathbb{R}^{12\times 12}$ are the state and control cost weight matrices. The simplified dynamics of the base is defined by $\mathbf{A}_j=\mathbf{A}\in\mathbb{R}^{12\times 12}$, $\mathbf{B}_j=\mathbf{B}\in\mathbb{R}^{12\times 12}$, and $\mathbf{b}_j=\mathbf{b}\in\mathbb{R}^{12\times 1}$ are constant at all time steps and are defined as:
\begin{equation}
    \begin{aligned}
        \mathbf{A}&=\begin{bmatrix}
            \mathbf{I}_{6\times 6}&\Delta t\mathbf{I}_{6\times 6}\\
            \mathbf{0}_{6\times 6}&\mathbf{I}_{6\times 6}
        \end{bmatrix},\mathbf{B}=\begin{bmatrix}
            \mathbf{B}^U\\\mathbf{B}^L
        \end{bmatrix},\mathbf{b}=\begin{bmatrix}
            \mathbf{0}_{3\times 3}\\\mathbf{0}_{3\times 3}\\
            \Delta t\mathbf{I}_{3\times 3}\\\mathbf{0}_{3\times 3}
        \end{bmatrix}\mathbf{g}\\
        \mathbf{B}^U&=\begin{bmatrix}
            \mathbf{0}_{6\times 6}&\mathbf{0}_{6\times 6}
        \end{bmatrix}\\
        \mathbf{B}^L&=\begin{bmatrix}
            \frac{\Delta t}{m}\mathbf{I}_{3\times 3}&...&\frac{\Delta t}{m}\mathbf{I}_{3\times 3}\\
            \Delta t\mathbf{I}_b^{-1}[\mathbf{r}_{{p_1}/b}]_k^{\times}&...&\Delta t\mathbf{I}_b^{-1}[\mathbf{r}_{{p_4}/b}]_k^{\times}
        \end{bmatrix}
    \end{aligned}
    \label{eq:mpc_dynamics}
\end{equation}
where $m$ is the mass of the robot, $\mathbf{I}_b\in\mathbb{R}^{3\times 3}$ is the total inertia of the robot in its nominal configuration, $\mathbf{r}_{p_i/b}=\mathbf{r}_{p_i,d,k}-\mathbf{r}_{b,d,k}\in\mathbb{R}^{3}$ is the relative position of the $i$-th foot with respect to the robot base. $[.]^{\times}\in so(3)$ represents the skew-symmetric matrix form for any vector $[.]\in\mathbb{R}^3$. The constraints on the control input consist of the unilaterality and friction pyramid constraints for each feet $i$. These are defined as:
\begin{equation}
    \begin{aligned}
        \mathbf{D}&=\begin{bmatrix}
            1&0&-\mu\\
            0&1&-\mu\\
            0&1&\mu\\
            1&0&\mu\\
            0&0&1
        \end{bmatrix},\\
        \mathbf{lb}_{i}&=\begin{bmatrix}
            -\infty\\
            -\infty\\
            0\\0\\\hat{\mathbf{s}}_{i,k}F_{z,min}
        \end{bmatrix},
        \mathbf{ub}_{i}=\begin{bmatrix}
            0\\0\\
            \infty\\
            \infty\\
            \hat{\mathbf{s}}_{i,k}F_{z,max}
        \end{bmatrix}
    \end{aligned}
    \label{eq:mpc_constraints}
\end{equation}
where $\mu$ is the friction coefficient, and $\hat{\mathbf{s}}_{i,k}$ is the current estimate of the contact state of the $i$-th foot. The contact estimates are provided by the contact detection algorithm used in \cite{bledt2018contact}.
The output from the MPC controller is sampled in the main control thread and fed to the reactive controller. The reactive controller enforces the full rigid-body dynamics and friction cone constraints, while maintaining a high rate of control commands update. The QP proposed in the original paper \cite{khandelwal2024distributed} is modified to the following:
\begin{equation}
    \begin{aligned}
    \mathbf{F}_{c,k}^* = \arg \min_{\mathbf{F}_c} \quad & (\mathbf{J}_{ab}^T\mathbf{F}_c-\boldsymbol{\tau}_b)^T\mathbf{S}_1(\mathbf{J}_{ab}^T\mathbf{F}_c-\boldsymbol{\tau}_b) \\
           & + \mathbf{F}_c^T\mathbf{W}\mathbf{F}_c \\
           & + (\mathbf{F}_c - \mathbf{F}^*_{c,mpc,k})^T\mathbf{V}(\mathbf{F}_c - \mathbf{F}^*_{c,mpc,k}) \\
    \text{subject to} \quad                        & \mathbf{g}(\mathbf{F}_c) \leq \mathbf{0}
    \end{aligned}
    \label{eq:didc_optimization}
\end{equation}
where $\mathbf{J}_{ab}^T\in\mathbb{R}^{6\times 12}$ is the sub-matrix of the Jacobian matrix $\mathbf{J}$ defined above, relating the contact forces on the feet to the base wrench. $\boldsymbol{\tau}_b\in\mathbb{R}^{6}$ is the desired feedback wrench on the robot base, $\mathbf{F}_{c,k}^*\in\mathbb{R}^{12}$ is the vector of the optimal contact forces on the feet that are solution to Eq. (\ref{eq:didc_optimization}). $\mathbf{S}_1\in\mathbb{R}^{6\times6}$, $\mathbf{W}\in\mathbb{R}^{12\times 12}$, and $\mathbf{V}\in\mathbb{R}^{12\times 12}$ are the relative weighing and regularization matrices for the respective terms. $\mathbf{F}^*_{c,mpc,k}$ is the contact force selected from the last MPC solution indexed at the current time-step. $\mathbf{g}(\mathbf{F}_c)$ is the vector representing the cone constraints on the contact forces for each leg.

To solve Eq. (\ref{eq:mpc_optimization}), we use the High Performance Interior Point Method (HPIPM) solver \cite{frison2020hpipm}. The optimization problem in the reactive control layer is solved using the Geometric Projected Gradient Descent (GPGD) algorithm \cite{khandelwal2024distributed}.

\section{Stability margin for gait frequency adaptation}
To maintain stability of the robot, it is not possible to use the conventional heuristics that utilize the projection of the center of mass positioned inside the contact polygon, since that polygon does not exist for dynamic gaits that have $\le2$ feet in contact with the ground. \cite{cheetah3} defined the concept of Virtual Predictive Support Polygon (VPSP) as an alternative to this for dynamic gaits. They used this dynamically changing polygon to generate the reference position of the robot base. Since this polygon is constantly changing in size, it cannot be used to reliably calculate a stability margin. In our experience the VPSP method makes it difficult for the robot to follow the desired base velocity since it always tries to re-place the horizontal coordinates of the base at the center of the polygon. To overcome these issues, we define a Time-Averaged Polygon for Stability (TAPS), that is calculated using the stance fraction, $\phi_{st}$, of the current gait and the estimated feet position, $\hat{\mathbf{r}}_{p_i,k}$. The vertices of this polygon are calculated as:
\begin{equation}
    \begin{aligned}
        \xi_{i,k}^{x,y}=\phi_{st}\hat{\mathbf{r}}_{{p_i},k}^{x,y}\forall i=\{1,2,3,4\}.
    \end{aligned}
\end{equation}
Since the vertices of this polygon are directly proportional to the gait stance fraction, this provides a metric to quantify the stability margin for different gaits, and only varies with changing foot position or stance fraction. This polygon is used to calculate the stability margin, $h$, to adapt the base height as well as gait frequency online as given in \textbf{Algorithm \ref{alg:gait_freq}}. The parameters $\eta^-$ and $\eta^+$ control how fast the gait period changes. Similarly, $\gamma^-$ and $\gamma^+$ control the rate of base height adjustment.
\begin{algorithm}[tp]
\caption{Gait Frequency and Base Height Adaptation}
\label{alg:gait_freq}
\begin{algorithmic}[1]
\Function{AdaptMotion}{ }
    \State $\xi_{i,k}\gets \phi_{st}\hat{\mathbf{r}}_{{p_i},k}$
    \State $h \gets \text{DistanceFromPolygon}(\xi^{x,y}_{i,k}, \hat{\mathbf{r}}_{b,k}^{x,y})$
    \If {$h < h_{min}$}
        \State $t_g = t_{g,max}$
        \State $r_{b,d,k}^z=(r_{b,d,k}^z)_{min}$
    \ElsIf {$h < h_{nom}$ or $d_{\text{limit}}$ is \textbf{true}}
        \State $t_g = t_g + \eta^+\Delta t$
        \State $r_{b,d,k}^z=r_{b,d,k}^z-\gamma^-\Delta t$
    \Else
        \State $t_g = t_g - \eta^-\Delta t$
        \State $r_{b,d,k}^z=r_{b,d,k}^z+\gamma^+\Delta t$
    \EndIf
    \State \Return $t_g$
\EndFunction
\end{algorithmic}
\end{algorithm}
Some of the advantages of this stability measure over the existing ones are: 1) it can be calculated for dynamic gaits, 2) its shape changes only with changing feet positions, and not with the instantaneous gait phase, 3) the dependence on gait through $\phi_{st}$ can be used to adjust the gait parameters of the robot if a larger stability region is required. This leads to better disturbance rejection and considerably shorter foot-step lengths during motion. It stabilizes the robot using only the natural evolution of the existing control inputs and does not need any additional corrective control inputs to stabilize the robot.

\section{EXPERIMENTS AND RESULTS}
\label{sec:results}
\subsection{Simulation Results}
The simulation environment used for these experiments was setup using the MuJoCo simulator. Additional features were added to the simulation to make it more realistic and reduce the sim-to-real gap. These features are: 1) control and sensor rate limiting, and 2) probabilistic latency modeling. For simulating the latency, the round trip time of the data to and from the robot was measured. A histogram was made of the measurements using bins of 5ms. On each simulation step, the latest control commands were pushed at the front of a double ended queue. The simulated latency was then used to calculate the index to be used in the control buffer to get the control commands for current time step. Higher latencies correspond to larger indices.
\subsubsection{Push-recovery}
\begin{figure}[htp]
    \centering
    \includegraphics[width=1.0\linewidth]{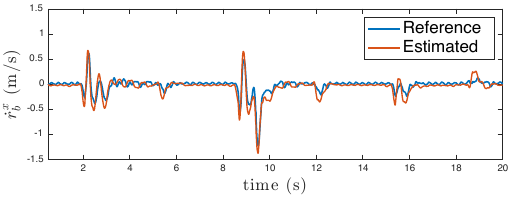}\\
    \includegraphics[width=1.0\linewidth]{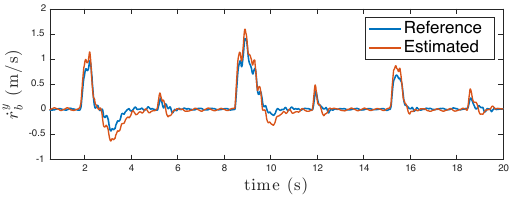}
    \caption{Velocity response of the robot in $x$ and $y$ axis after push along +$y$ axis (simulation).}
    \label{fig:perturb_base_vel}
\end{figure}

To quantify improvement in the push-recovery, the robot was subjected to impulsive pushes inside the simulation. The magnitude of the impulse was chosen to be $50000$ N/sec and it was applied along the $^\mathcal{G}\hat{y}$ axis. The push was applied at a frequency of $0.3$ Hz. The robot was able to maintain its state for a disturbance $\sim1.5$m/s in the $^\mathcal{G}\hat{y}$ direction. The evolution of the robot base velocity during the experiment is shown in Figure \ref{fig:perturb_base_vel}.

\begin{figure}[htp]
    \centering
    \includegraphics[width=1.0\linewidth]{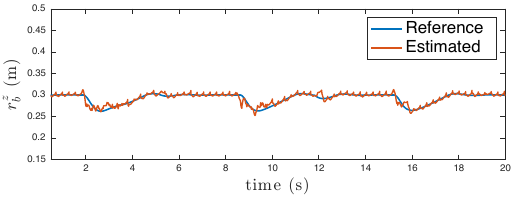}
    \caption{Base height adjustment during push experiment (simulation).}
    \label{fig:base_height_adjustment}
\end{figure}

Figure \ref{fig:base_height_adjustment} shows the base height adjustment for this period, showing that the proposed stability margin measure, TAPS, is leading to base height adjustment only when the robot is destabilized by external forces.

The difference in response for the same push is because the stabilization of the robot is dependent on the instantaneous gait phase. The swing legs can travel farther if the push is applied at the start of the swing phase since the robot feet can travel farther at moderate velocities to mitigate the disturbance, hence requiring less foot contact forces. In contrast, the robot relies almost entirely on the foot contact forces for stabilization if the disturbance is applied near or at the end of the swing phase. During this time, the swing feet also have to travel farther at higher velocity, increasing the chances of foot slip.

\subsubsection{Assistive load carrying}
\begin{figure}[htp]
    \centering
    \includegraphics[width=1.0\linewidth]{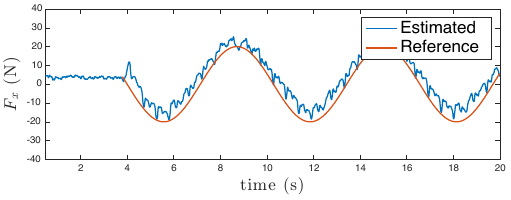}\\
    \includegraphics[width=1.0\linewidth]{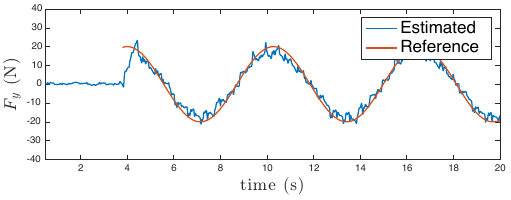}
    \caption{Applied and estimated force on the robot (simulation).}
    \label{fig:perturb_circle_fx_fy}
\end{figure}

\begin{figure}[htp]
    \centering
    \includegraphics[width=1.0\linewidth]{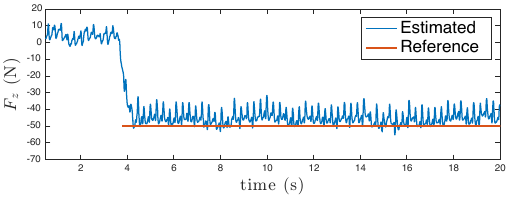}
    \caption{Vertical base load estimation comparison for $F_z=50$ N (simulation).}
    \label{fig:base_load}
\end{figure}

\begin{figure}[htp]
    \centering
    \includegraphics[width=1.0\linewidth]{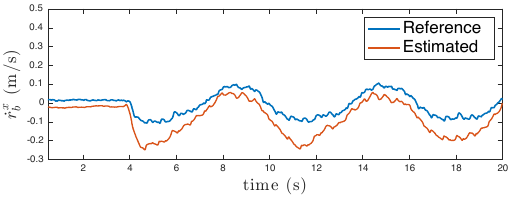}\\
    \includegraphics[width=1.0\linewidth]{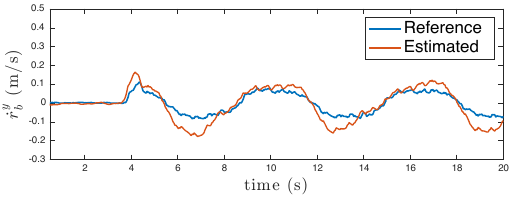}
    \caption{Base velocity for the applied circular base force (simulation).}
    \label{fig:perturb_circle_base_vel}
\end{figure}

In simulation, a constant vertical force of $F_z=-50$ N was applied to the robot. The robot was then guided by a horizontal force of magnitude $|F|=20$ N varying circularly with $\omega=1$ rad/sec. 
\begin{equation}
    \begin{aligned}
        F_{ref}^{x,y}=|F|(-\sin(\omega t)\hat{x}+\cos(\omega t)\hat{y}) .
    \end{aligned}
\end{equation}
There is a delay of $\sim 4$ seconds between the time at which recording of the data was started and the time when forces were first applied in the simulation, as is visible in the figures. The base force estimation algorithm is able to give accurate estimates of the horizontal time-varying forces, as shown in Figure \ref{fig:perturb_circle_fx_fy}, as well as the constant vertical load on the body, as shown in Figure \ref{fig:base_load} . The velocity response due to the applied force is shown in Figure \ref{fig:perturb_circle_base_vel}.

\subsection{Experimental results on Go2 quadruped robot}
\begin{figure}[htp]
    \centering
    \includegraphics[width=0.7\linewidth]{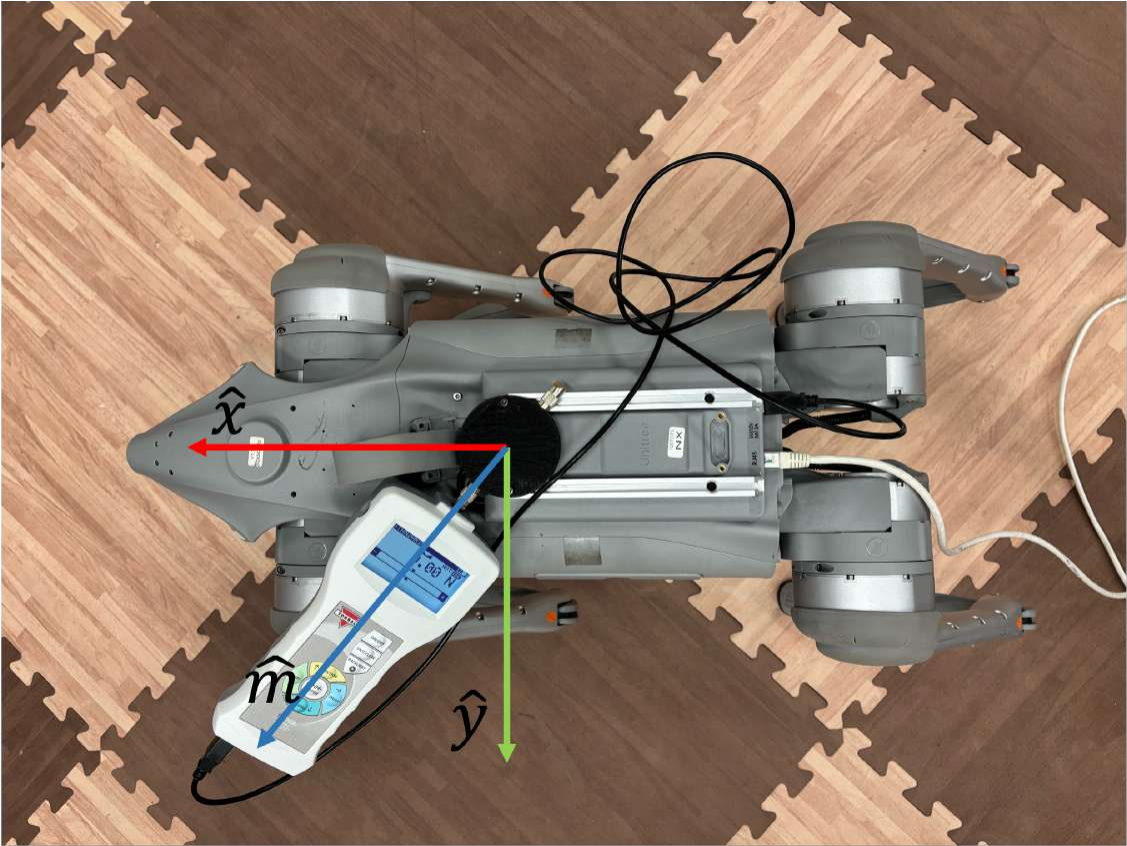}
    \caption{Ground truth force measurement setup on the robot.}
    \label{fig:fb200_go2}
\end{figure}

The proposed controller was tested and validated on the Go2 robot from Unitree Robotics. The software stack used in \cite{khandelwal2024distributed} was extended for this work. The control commands were calculated and sent to the low-level control API using the on-board Nvidia Jetson Orin-NX with a Arm Cortex-A78AE CPU. CycloneDDS \cite{cyclonedds} was used as the preferred communication middleware to send the commands to the robot. To measure the ground truth force applied on the robot base, the FB200 single-axis force sensor was mounted on the robot base, shown in Figure \ref{fig:fb200_go2}. The force sensor data was sent via USB at 80Hz to the Orin-NX board, where it was decoded using the pyserial library and published as a ROS2 topic (with CycloneDDS middleware) to allow recording alongside all the other data in a ros2 bag.

\subsubsection{No load pushing/pulling}
\begin{figure}[htp]
    \centering
    \includegraphics[width=1.0\linewidth]{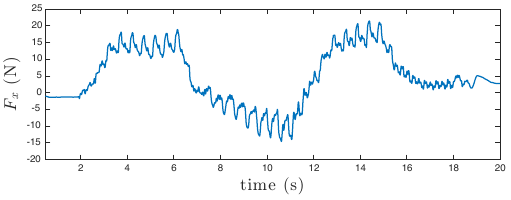}\\
    \includegraphics[width=1.0\linewidth]{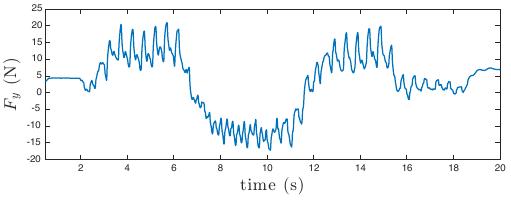}
    \caption{Estimated forces along the $x$ and $y$ axis on the robot base (hardware).}
    \label{fig:fx_fy}
\end{figure}

\begin{figure}[htp]
    \centering
    \includegraphics[width=1.0\linewidth]{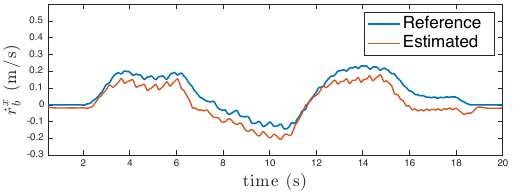}\\
    \includegraphics[width=1.0\linewidth]{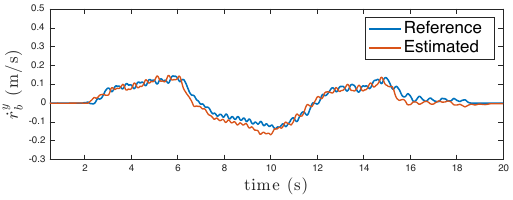}\\
    \caption{Base velocity response due to applied force (hardware).}
    \label{fig:base_vel}
\end{figure}

\begin{figure}[htp]
    \centering
    \includegraphics[width=1.0\linewidth]{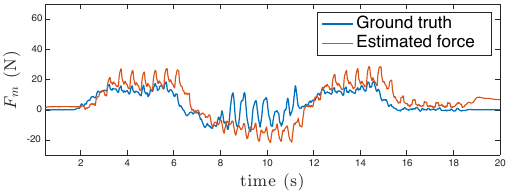}
    \caption{Comparison of measured force and projection of measured force along the measurement direction (hardware).}
    \label{fig:fn_compare}
\end{figure}

As mentioned previously, a single-axis force sensor (FB200) was mounted on the robot base. The force sensor was mounted on the base of the robot along the measurement axis $^\mathcal{G}\hat{m}=$$^\mathcal{G}\hat{x} +$$^\mathcal{G}\hat{y}$, and the robot was pulled and pushed along this axis.
The estimated base forces along the $^\mathcal{G}\hat{x}$ and $^\mathcal{G}\hat{y}$ axes are shown in Figure \ref{fig:fx_fy}.
The response of the base velocity due to the applied force is shown in Figure \ref{fig:base_vel}. Even with no joystick input, the reference velocity changes due to the applied force, which moves the robot in the same direction ($\hat{m}$ in this case) as the applied force.
To establish the accuracy and validity of the estimated base wrench, the estimated forces from the base wrench estimates are checked against the ground truth data obtained using the mounted force sensor. Figure \ref{fig:fn_compare} shows the comparison of the measured force from the sensor along with the projection of estimated base force along the measurement direction ($^\mathcal{G}\hat{m})$. This is denoted as $F_m$ in the figure. Note that both the signals have been filtered using a windowed moving average filter with different window sizes to show the evolution of forces on average. This is one of the reasons for the phase lag visible in the figure.
There is a mismatch in the estimated forces and the ground truth between time $t=8-12$ seconds in Figure \ref{fig:fn_compare}. This is due to the human error made while trying to align the manually applied external forces along the measurement axis. 
\subsubsection{Push recovery}
\begin{figure}[htp]
    \centering
    \includegraphics[width=1.0\linewidth]{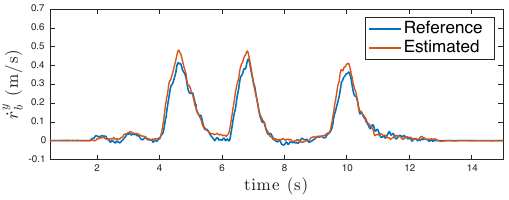}\\
    \includegraphics[width=1.0\linewidth]{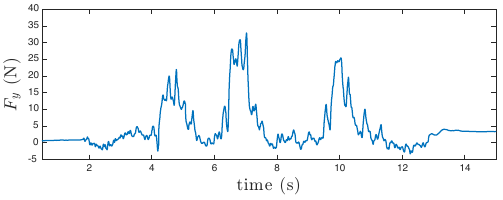}
    \caption{Velocity response of the robot after push along $\hat{y}_b$ axis (hardware). The bottom plot shows the estimated force on the base.}
    \label{fig:push_hw_vy}
\end{figure}

To test the push-recovery behavior, the robot was pushed manually in the $^\mathcal{G}\hat{y}$ direction. Figure \ref{fig:push_hw_vy} shows the velocity response of the robot for multiple pushes. The corresponding force estimates are shown in the same figure.

\subsubsection{Assistive load carrying}
\begin{figure}[htp]
    \centering
    \includegraphics[width=1.0\linewidth]{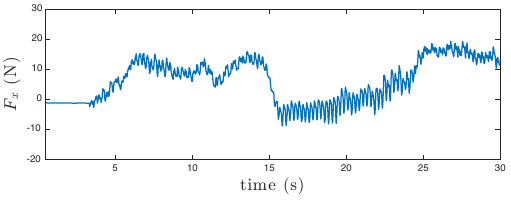}\\
    \includegraphics[width=1.0\linewidth]{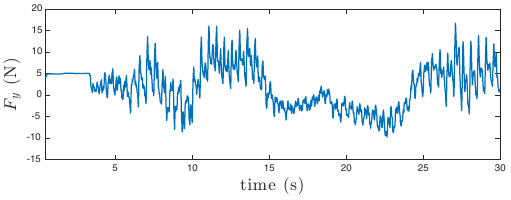}\\
    \includegraphics[width=1.0\linewidth]{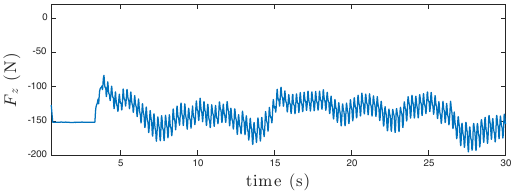}
    \caption{Estimated base forces during assistive load carrying experiment (hardware).}
    \label{fig:base_loaded_forces}
\end{figure}

\begin{figure}[htp]
    \centering
    \includegraphics[width=1.0\linewidth]{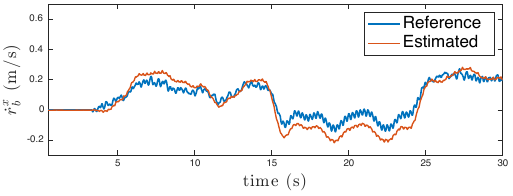}\\
    \includegraphics[width=1.0\linewidth]{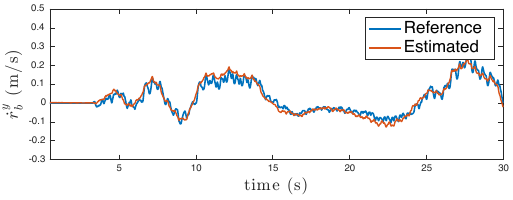}
    \caption{Velocity response during the assistive load carrying experiment (hardware).}
    \label{fig:base_loaded_vel}
\end{figure}

For this experiment, a load of $\sim150$N was mounted on the robot. The robot was then pushed around manually in varying directions. The estimated base forces are shown in Figure \ref{fig:base_loaded_forces}.
The resulting velocity response in the horizontal plane is shown in Figure \ref{fig:base_loaded_vel}.
\subsubsection{CBF based collision avoidance}
\begin{figure}[htp]
    \centering
    \includegraphics[width=1.0\linewidth]{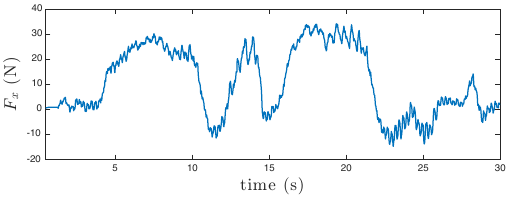}\\
    \includegraphics[width=1.0\linewidth]{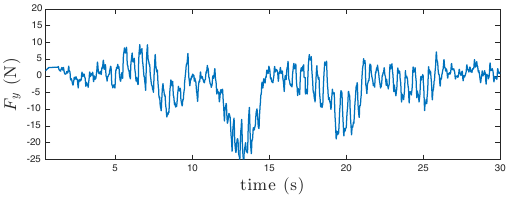}
    \caption{Estimated base forces during the collision avoidance test (hardware).}
    \label{fig:cbf_force}
\end{figure}

\begin{figure}[htp]
    \centering
    \includegraphics[width=1.0\linewidth]{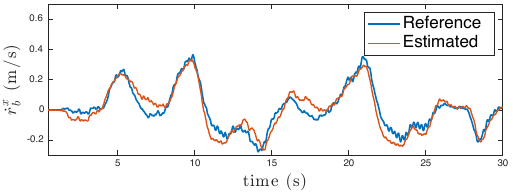}\\
    \includegraphics[width=1.0\linewidth]{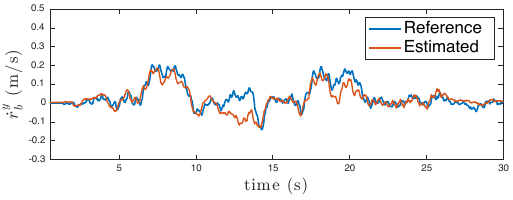}\\
    \caption{Velocity response of the robot base during collision avoidance test (hardware).}
    \label{fig:cbf_vel}
\end{figure}

Since we are using CBF to avoid bumping into the leader robot, we get collision avoidance automatically, only in front of the robot. To test the proposed CBF based avoidance, a cuboidal cardboard box was placed in front of the robot. The robot was then pushed manually towards one of the edges on the front face of the box. Figure \ref{fig:cbf_force} show the estimated forces applied on the robot base during this experiment. Figure \ref{fig:cbf_vel} shows the reference and estimated velocity of the robot during this experiment. The parameters chosen for the CBF were: $\alpha=500$, $T_h=0.5$, $\delta_{CBF}=0.3$.
It can be observed that even when the robot was pushed directly towards the box, the CBF based planner changed the reference, and hence estimated, base velocities. Note the velocity response of the robot in $^\mathcal{G}\hat{x}$ direction between 5-10 second period. A significant external force ($\sim30$N) was applied on the robot along $^\mathcal{G}\hat{x}$ axis, but due to the presence of the box it did not move in the $^\mathcal{G}\hat{x}$ axis and instead started moving along $^\mathcal{G}\hat{y}$ axis even when no external force was being applied on the robot in that direction. This shows that the proposed collision avoidance pipeline is working as expected in conjunction with the admittance control.
\subsubsection{Human-robot collaborative load carrying}
\begin{figure}[htp]
    \centering
    \includegraphics[width=0.8\linewidth]{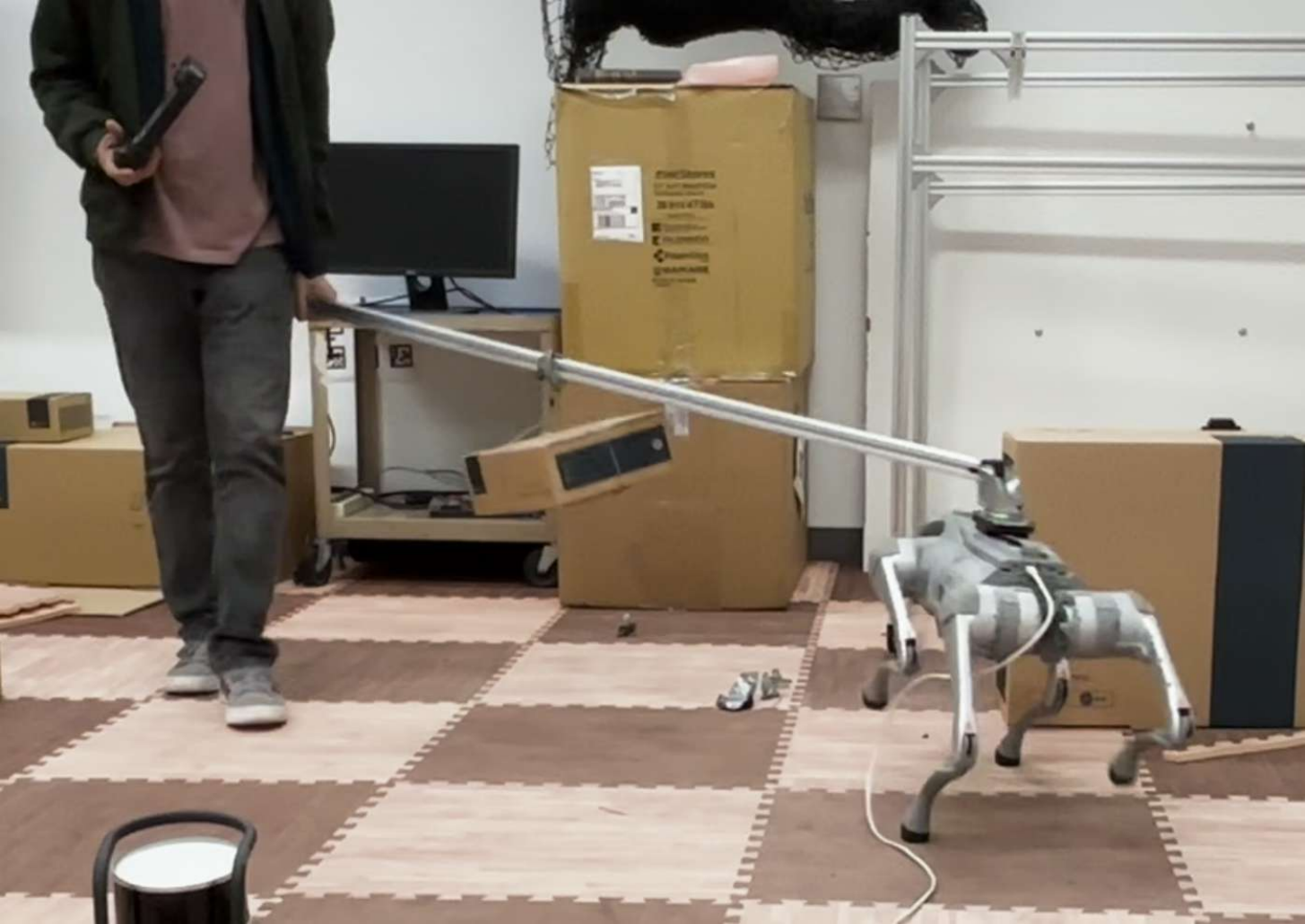}
    \caption{Human-robot collaborative load carrying experiment.}
    \label{fig:human_robot_collab}
\end{figure}

\begin{figure}[htp]
    \centering
    \includegraphics[width=1.0\linewidth]{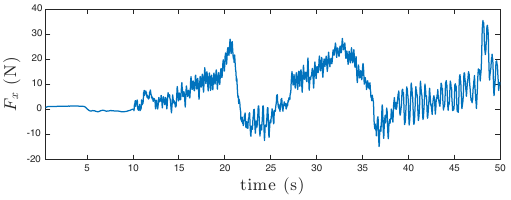}\\
    \includegraphics[width=1.0\linewidth]{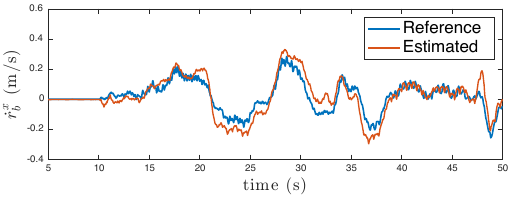}
    \caption{Estimated base force and the resulting base velocity along the x-axis during the human-robot collaborative load carrying experiment.}
    \label{fig:fx_vx_human_robot_collab}
\end{figure}

\begin{figure}[htp]
    \centering
    \includegraphics[width=1.0\linewidth]{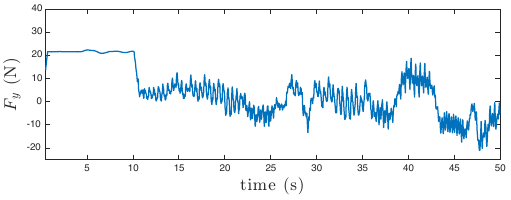}\\
    \includegraphics[width=1.0\linewidth]{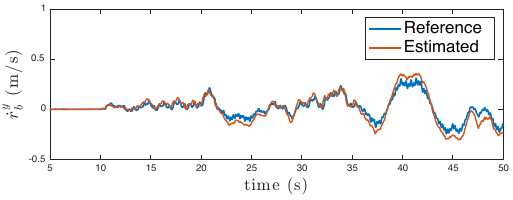}
    \caption{Estimated base force and the resulting base velocity along the y-axis during the human-robot collaborative load carrying experiment.}
    \label{fig:fy_vy_human_robot_collab}
\end{figure}

In this experiment, a load of $\sim120$N was mounted on a rigid shaft connected to the robot base with a spherical joint at one end. The other end was supported by the human operator and the robot was then moved around by applying forces to the other end of the rigid rod. Figure \ref{fig:human_robot_collab} shows the setup of the experiment from one time instant during the experiment. 
The resulting base forces and the velocity response(s) along $^\mathcal{G}\hat{x}$ and $^\mathcal{G}\hat{y}$ axis are shown in Figure \ref{fig:fx_vx_human_robot_collab} and Figure \ref{fig:fy_vy_human_robot_collab} respectively. The initial pre-load visible in estimated force in Figure \ref{fig:fy_vy_human_robot_collab} is the force on the robot resulting due to unbalanced force in $^\mathcal{G}\hat{y}$ axis when the robot was in stance phase and the feedback controller is not engaged. This is compensated when the robot changes to trot gait and the feedback controller becomes active, as is visible in the figure around time $t=10$ seconds.
\subsubsection{Robot-robot collaborative load carrying}
\begin{figure}[htp]
    \centering
    \includegraphics[width=0.8\linewidth]{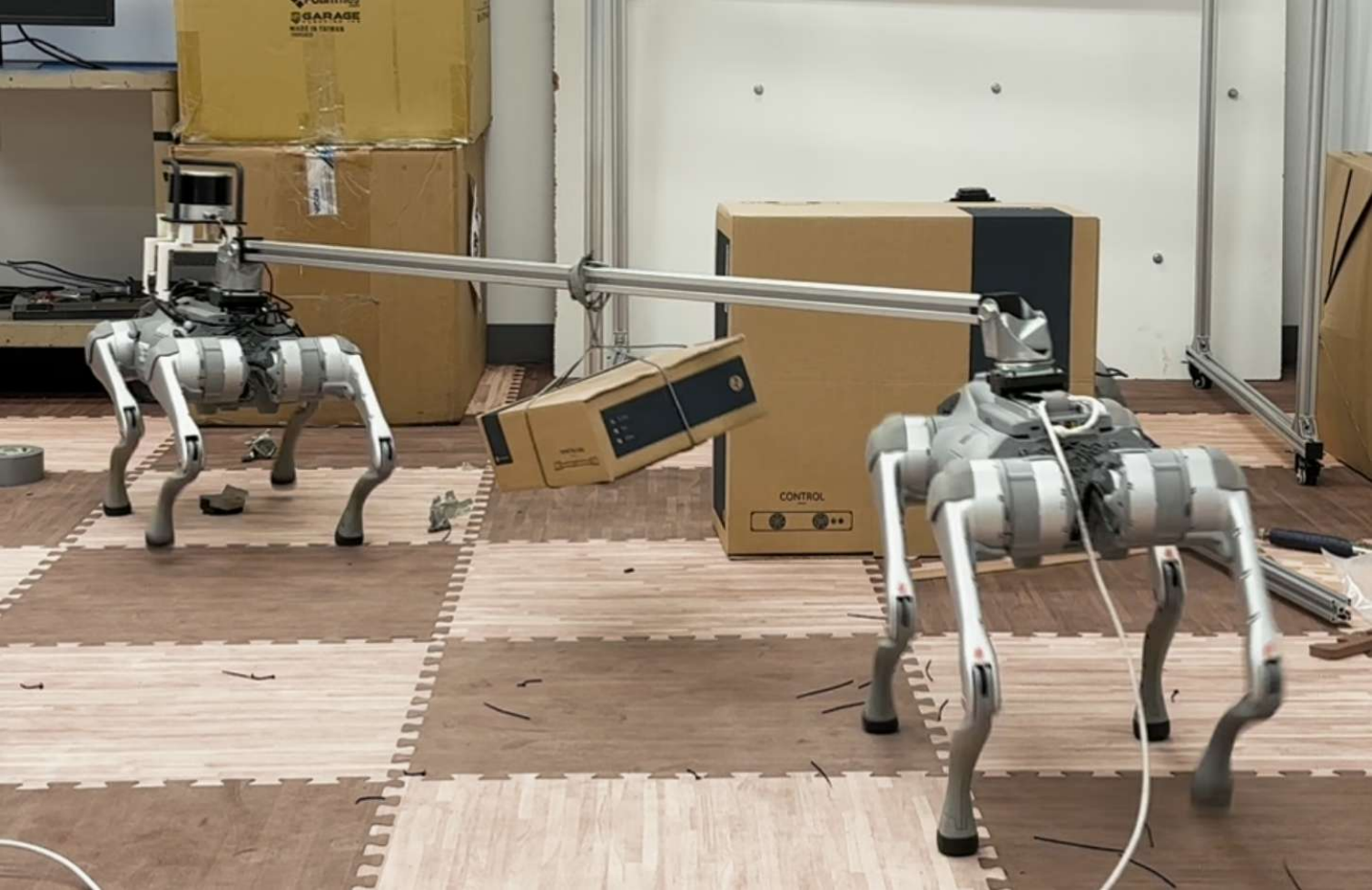}
    \caption{Robot-robot collaborative load carrying experiment.}
    \label{fig:robot_robot_collab}
\end{figure}

\begin{figure}[htp]
    \centering
    \includegraphics[width=1.0\linewidth]{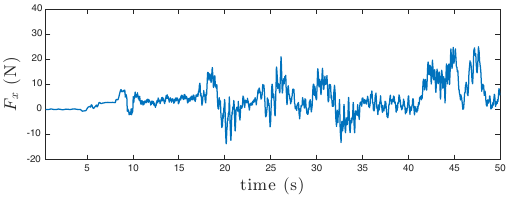}\\
    \includegraphics[width=1.0\linewidth]{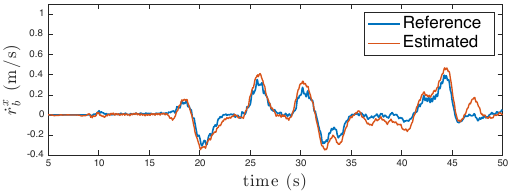}
    \caption{Estimated base force and the resulting base velocity along the x-axis during the robot-robot collaborative load carrying experiment.}
    \label{fig:fx_vx_robot_robot_collab}
\end{figure}

\begin{figure}[htp]
    \centering
    \includegraphics[width=1.0\linewidth]{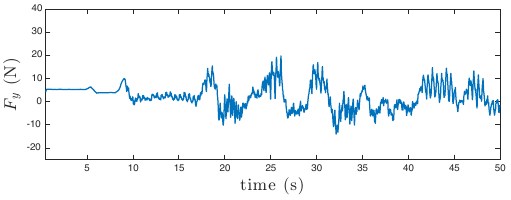}\\
    \includegraphics[width=1.0\linewidth]{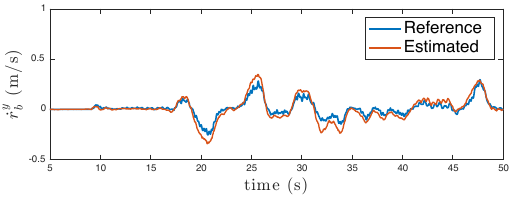}
    \caption{Estimated base force and the resulting base velocity along the y-axis during the robot-robot collaborative load carrying experiment.}
    \label{fig:fy_vy_robot_robot_collab}
\end{figure}

The same setup from the previous experiment was tested for robot-robot load carrying, where in place of the human operator another quadruped robot (called the \textit{leader} robot) was used to move the guiding end of the connecting rod. The leader robot was controller using joystick by a human operator and the other robot running the proposed controller was allowed to move freely based on the forces applied on it via the rigid rod connecting the two robots. This setup is shown in Figure \ref{fig:robot_robot_collab} where the leader robot is pulling the robot via the rod and the other robot is moving as guided by it. For the leader robot, the internal motion control algorithm provided by the robot vendor was used.
The estimated base forces applied to the follower robot by the leader robot through the rod and the corresponding velocity response along $^\mathcal{G}\hat{x}$ and $^\mathcal{G}\hat{y}$ axis are shown in Figure \ref{fig:fx_vx_robot_robot_collab} and Figure \ref{fig:fy_vy_robot_robot_collab} respectively. During the experiment, both the admittance controller and the CBF based collision avoidance is active. This is visible in the figures as well when the direction of the velocity response does not match the direction of the estimated applied force. This experiment also shows that the proposed framework is able to work for any generic follower that can apply forces to the guiding end of the connecting rod, and thus does not need any modifications to work with different leader agents (humans, quadrupeds, humanoid, wheeled robots, etc.).
\section{CONCLUSION}
We proposed a methodology for designing a compliant controller for the quadruped robot with application to assistive load carrying using admittance control. The proprioceptive data from the robot was used to estimate the external forces required for the admittance control. The motion planner for the robot decouples the base motion planning and foot motion planning with a heuristic footstep placement strategy. The base motion, since decoupled, was used to perform second order (acceleration) control of the robot base. The single acceleration command was used to track the joystick commands, ensure compliant behavior using admittance control, and avoid collisions using CBFs. Multiple experiments were carried out to test and validate the proposed control and planning pipeline in simulation and on physical hardware, including the human-robot and robot-robot collaborative load carrying.







\bibliographystyle{IEEEtran}
\bibliography{references.bib}

\end{document}